# An Eigenspace Divide-and-Conquer Approach for Large-Scale Optimization

Zhigang Ren, Yongsheng Liang, Muyi Wang, Yang Yang, and An Chen

School of Automation Science and Engineering, Xi'an Jiaotong University, Xi'an, China

**Corresponding author:** Zhigang Ren

**Affiliation:** School of Automation Science and Engineering, Xi'an Jiaotong University

**Permanent address:** No.28 Xianning West Road, Xi'an, Shaanxi, 710049, P.R. China

**Email address:** renzg@mail.xjtu.edu.cn

# An Eigenspace Divide-and-Conquer Approach for Large-Scale Optimization

Zhigang Ren, Yongsheng Liang, Muyi Wang, Yang Yang, and An Chen

School of Automation Science and Engineering, Xi'an Jiaotong University, Xi'an, China

Abstract:

Divide-and-conquer-based (DC-based) evolutionary algorithms (EAs) have achieved notable success in dealing with large-scale optimization problems (LSOPs). However, the appealing performance of this type of algorithms generally requires a high-precision decomposition of the optimization problem, which is still a challenging task for existing decomposition methods. This study attempts to address the above issue from a different perspective and proposes an eigenspace divide-and-conquer (EDC) approach. Different from existing DC-based algorithms that perform decomposition and optimization in the original decision space, EDC first establishes an eigenspace by conducting singular value decomposition on a set of high-quality solutions selected from recent generations. Then it transforms the optimization problem into the eigenspace, and thus significantly weakens the dependencies among the corresponding eigenvariables. Accordingly, these eigenvariables can be efficiently grouped by a simple random strategy and each of the resulting subproblems can be addressed more easily by a traditional EA. To verify the efficiency of EDC, comprehensive experimental studies were conducted on two sets of benchmark functions. Experimental results indicate that EDC is robust to its parameters and has good scalability to the problem dimension. The comparison with several state-of-the-art algorithms further confirms that EDC is pretty competitive and performs better on complicated LSOPs.

Keywords: large-scale optimization; eigenspace divide-and-conquer; space transformation; variable dependency

## 1. Introduction

With the fast development of big data technologies, many real-world optimization problems show an enormous increase in the number of decision variables, and large-scale optimization has become an active research field in the last decade. Lots of science and engineering applications, such as large-scale flexible scheduling [1], gene array classification [2], satellite-module layout design [3] and overlapping community detection in large-scale networks [4], can be formulated as large-scale optimization problems (LSOPs). However, now it is still a very challenging task to solve LSOPs. It has been broadly recognized that the performance of traditional evolutionary algorithms (EAs) would significantly deteriorate when dealing with LSOPs [5,6]. The reason mainly lies in that the solution space of a problem exponentially increases with the growth of its dimension and it is non-trivial for a traditional EA to adequately explore a huge solution space within acceptable computation time. This is the so-called "curse of dimensionality" [7].

Large-scale optimization has attracted lots of research efforts and motivated the development of a great number of advanced optimization algorithms in recent years [5,8,9,10]. Generally, these algorithms can be classified into two major categories, namely, divide-and-conquer-based (DC-based) methods and non-DC-based ones. Non-DC-based methods directly tackle an LSOP in the whole solution space. They are generally accompanied with specifically designed operators or combined with different kinds of efficient search techniques such that their search capability in high-dimensional space can be enhanced. Representatives of this type of methods include the famous MA-SW-Chains [11] and multiple-offspring-sampling-based (MOS-based) algorithm [12]. In contrast, DC-based methods address LSOPs in a divide-and-conquer manner. They first decompose an LSOP into several lower-dimensional subproblems and then cooperatively optimize them with some traditional optimizers. Cooperative coevolution (CC) [9,13] is one of the most important parts of DC-based methods.

DC-based methods show several advantages over non-DC-based ones. Firstly, they provide an effectual way to alleviate the "curse of dimensionality" by decomposing an LSOP into multiple subproblems of small sizes. Secondly, they tackle different subproblems with different subpopulations, which is beneficial to increase the solution diversity and improve their robustness [14,15]. Finally, they support parallel optimization on subproblems, which can greatly speed up the whole optimization process. On the other side, DC-based methods also present some limitations. They seldom achieve satisfying performance unless the problem in hand is properly decomposed [13,16]. If some interacting variables are improperly partitioned into different subproblems, DC-based methods generally lose their efficiency and the optimizers therein tend to end up at Nash equilibria rather than the true optima [17]. Unfortunately, now it is still non-trivial to properly decompose an LSOP [9,18], not to mention that many real-world optimization problems are nonseparable in a strict sense [19,20].

From the above analysis, it can be known that problem decomposition and subproblem optimization are two important algorithmic components in DC-based methods, and the former provides a foundation for the latter. Although the concrete implementations of different existing DC-based methods are diverse, one of their common characteristics is that they all conduct decomposition and optimization in the original solution space. The dependencies among the original decision variables of some LSOPs are so complicated that they pose a great challenge to both the decomposer and the optimizer. If the variable dependencies can be weakened, the resulting problem will become more tractable.

In this study, we attempt to address the above issue with space transformation techniques [21,22]. Space transformation is an effective tool to reduce variable dependencies, but has seldom been employed to tackle LSOPs. By integrating the advantages of a space transformation technique and a DC strategy together, this study develops an eigenspace divide-and-conquer (EDC) approach for large-scale optimization. More specifically, EDC contains the following three algorithmic components:

1) Space transformation. When tackling an LSOP, EDC first builds an eigenspace by performing singular value decomposition (SVD) on some high-quality solutions, and then transforms the original population into the established eigenspace such that the transformed population can be partitioned and evolved more easily. This mainly profits from the weak dependencies among the corresponding eigenvariables. After evolutionary operations, EDC transforms the obtained offspring population back into the original solution space for evaluation since the fitness function of the problem is defined there. The eigenspace is updated occasionally rather than in each generation, and thus brings no significant computation burden to the whole algorithm.

2) Eigenspace decomposition. Through space transformation, the original LSOP is converted to a new problem with many weakly dependent variables in the eigenspace. Benefiting from this characteristic, it is reasonable to randomly divide these eigenvariables into multiple disjoint subgroups and partition the transformed population into several subpopulations according to the grouping results. The employed random decomposition strategy can be easily executed without consuming any time-consuming fitness evaluations. Besides, it is also conducive to improving the population diversity.

3) Subproblem optimization. EDC adopts an efficient estimation of distribution algorithm (EDA) to separately and concurrently evolve all the subpopulations in the eigenspace. It neglects the weak interactions among different subproblems, but still considers the dependencies among variables in the same subproblem. Accordingly, the whole search space can be significantly reduced and a fine search within each low-dimensional eigensubspace can be achieved.

To verify the efficiency of the proposed EDC approach, this study tested it on two sets of benchmark functions whose dimensions vary from 100 to 1000 [6,23]. Experimental results demonstrate that EDC is robust to its parameters and has good scalability to the problem dimension. The combination of the space transformation and the DC strategy endows EDC the capability of efficiently solving a broad array of complicated LSOPs. Comparison with a number of state-of-the-art algorithms further confirms its strong competitiveness.

The remainder of this paper is organized as follows. Section 2 reviews the optimization algorithms for large-scale optimization, including the DC-based methods and non-DC-based ones. Section 3 elaborates the idea and implementation of the proposed EDC approach. Then in Section 4, the efficiency of EDC is comprehensively studied. Finally, conclusions and future work are given in Section 5.

## 2. Literature review
### 2.1 DC-based methods

Plenty of DC-based methods have been developed for large-scale optimization, where CC algorithms are a class of representatives.

1) CC algorithms. CC first decomposes an LSOP into a set of lower-dimensional subproblems and then cooperatively optimizes them with a traditional EA in an iterative manner. A key feature of CC is that it only optimizes a single subproblem at a time and evaluates each subsolution based on a context vector, which is a complete solution composed of representative subsolutions [9,13].

Problem decomposition is a fundamental algorithmic component in CC. The first CC algorithm named cooperative coevolutionary genetic algorithm [13] decomposes an $n$-dimensional problem into $n$ 1-dimensional subproblems, and was shown to work well on problems with no interacting variables. Bergh and Engelbrecht [24] suggested decomposing an $n$-dimensional problem into $m$ $s$-dimensional subproblems, where $n=m \times s$, and showed that this strategy performs better than the 1-dimensional decomposition approach on a variety of problems. However, the above static decomposition methods can hardly assign interacting variables into the same subcomponent and show performance deterioration on nonseparable problems. To alleviate this limitation, Yang et al. [25] proposed a random decomposition method, which randomly redivides all the variables into different subcomponents in each CC cycle. Recently, some learning-based decomposition methods [16,26,27,28] were developed and got high research attention. They can achieve desirable decomposition results by explicitly detecting variable dependencies at the cost of sampling and evaluating a considerable number of solutions. Nevertheless, as variable interactions becoming more and more complicated, existing decomposition methods generally lose their efficiency and cannot guarantee the correctness of the decomposition, which would severely limit the performance of CC. How to design efficient and widely applicable decomposition strategies remains an open issue in CC [9].

After decomposition, CC sequentially optimizes all the subproblems with a traditional EA in each cycle. So far, many kinds of EAs have been integrated into CC such as genetic algorithm (GA) [13,29], evolution strategy (ES) [30], particle swarm optimization (PSO) [31], differential evolution (DE) [26] and EDA [32]. During the optimization process, all the subproblems cooperate with each other by providing the context vector for subsolution evaluation. Canonical CC algorithms usually employ the best solution found so far as the context vector [16,31], which may cause CC converging to Nash equilibria rather than real optima if the original problem is improperly decomposed [33]. To alleviate this issue, some researchers suggested using single random context vector [34,35] or multiple context vectors [36,37] to achieve a more robust subsolution evaluation. For example, Peng et al. proposed a multimodal optimization enhanced CC (MMO-CC) [37], which constructs multiple context vectors by applying a nondominance-based selection strategy to the multiple optima concurrently achieved on each subproblem. Besides the cooperative relationship, subproblems in CC also compete with each other for limited computation resources (CRs). Canonical CC algorithms equally allocate CRs among subproblems, which is unreasonable because subproblems are generally imbalanced in terms of dimension, solving difficulty, and contribution to the total fitness [38]. Directing against this issue, contribution-based CC algorithms [38,39] dynamically allocate CRs in each cycle according to subproblems' contributions to the total fitness improvement. For instance, Ren et al. [40] recently proposed an efficient fine-grained CR allocation strategy. It takes a single iteration as an allocation unit and always allocates CRs to the subproblem that is most likely to make the largest overall contribution. Moreover, some other researchers attempted to make better use of CRs by exploiting solutions already evaluated. They constructed surrogate models for each subproblem with these solutions and achieved approximate and quick evaluation of candidate subsolutions by virtue of the resulting surrogate models [41,42]. As a result, many time-consuming real fitness evaluations can be avoided. However, now it is difficult to train an accurate enough surrogate model, especially when a subproblem is still of large scale.

2) Other DC-based algorithms. In addition to CC, there are also some other DC-based optimization algorithms. Dong et al. [43] proposed an EDA with a model complexity control (EDA-MCC) technique. EDA-MCC first identifies weakly dependent variables according to the linear correlation coefficient, and then forcibly divides the rest strongly dependent variables into several non-overlapping subsets with the aim of reducing the search space. It concurrently evolves all the variable subsets with different probabilistic models and shows superior performance over traditional EDAs and some other efficient algorithms on a set of benchmark functions. Xu et al. [44] further improved EDA-MCC by replacing the linear correlation coefficient with mutual information such that the nonlinear dependencies among variables can be detected. Yang et al. [18] proposed a self-evaluation evolution (SEE) algorithm by combing the DC strategy with the surrogate model technique. SEE divides an $n$-dimensional problem into $n$ 1-dimensional subproblems and employs a local search operator to separately evolve each subproblem. Efficient subsolution evaluation is achieved by training a simple surrogate model for each pair of offspring and parent subsolutions. Recently, Yang et al. [45] further developed an efficient algorithm named naturally parallelizable divide-and-conquer (NPDC) by enhancing SEE with a parallel computing technique.

**2.2 Non-DC-based methods**

Non-DC-based methods deal with the original LSOP as a whole. A common way is to integrate existing EAs with different kinds of search techniques to improve their search efficiency in the high-dimensional space. Two representatives are MA-SW-Chains [11] and MOS [12]. The former creates local search chains based on Solis-Wets method and employs them to exploit promising solution regions found by a GA, while the latter combines the Solis-Wets algorithm and the first of the local searches of the multiple trajectory search algorithm within the MOS framework. MA-SW-Chains and MOS won the CEC'2010 and CEC'2013 competitions on large-scale optimization, respectively, and are still state-of-the-art now.

Traditional EAs could be scaled up to solve LSOPs with some other alterations. Li and Zhang [46] suggested a rank one ES and a rank-$m$ ES (Rm-ES) for large-scale optimization by using sparse and low rank model in ES. Yang et al. [47] proposed a segment-based predominant learning swarm optimizer (SPLSO), which randomly divides all the variables into different segments and separately and concurrently evolves variables in terms of the segments. Yildiz et al. [48] developed a micro DE variant with a directional local search algorithm. For sake of maintaining a good balance between exploration and exploitation, it performs exploitation through DE operators and the directional local search and ensures exploration by randomly reinitializing the worst solutions to diverse solution regions.

Besides, it is also interesting to narrow the search space using dimensionality reduction techniques. Kabán et al. [49] developed a random-projection-based EDA (RP-EDA), which projects the original solution space into many low-dimensional spaces and employs a traditional EDA to find promising solutions in each low-dimensional space. Theoretical and experimental studies both verified the effectiveness of RP-EDA. Wang et al. [50] utilized random embeddings to simplify the original high-dimensional problem into a low-dimensional problem that only contains the most important variables. They further combined random embeddings with a Bayesian optimization algorithm, and showed that the resultant method can effectively solve a variety of high-dimensional problems. A comprehensive review of non-DC-based methods for large-scale optimization can be found in [5].

**3. Eigenspace Divide-and-Conquer**

The framework of EDC is presented in Fig. 1. Firstly, EDC generates an initial population in the original space and computes an eigen coordinate system to establish an eigenspace. Then EDC transforms the population into the eigenspace using a space transformation technique. In the eigenspace, the transformed population is divided into multiple subpopulations, which are separately and concurrently evolved by an optimizer. The resulting offspring subpopulations are merged together to generate a complete offspring population, which is finally transformed back into the original space for evaluation and selection. EDC executes an iterative process of the above operations until a termination condition is satisfied.

In the following subsections, the three main components of EDC, including space transformation, eigenspace decomposition and subproblem optimization, will be described in detail.

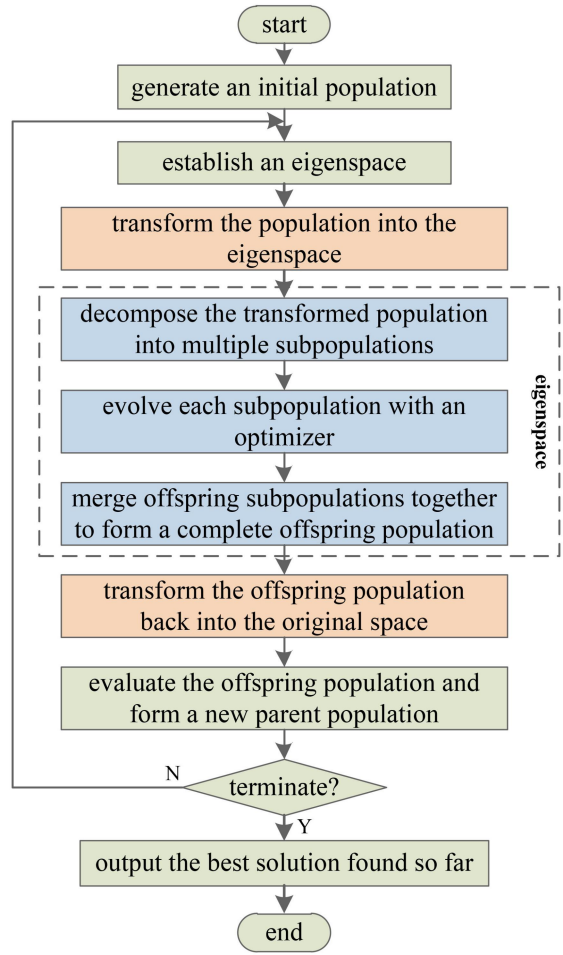

Figure 1. The framework of EDC

**3.1 Space transformation**

EDC utilizes space transformation techniques to handle variable dependencies. A desirable transformation should satisfy two major requirements: First, the transformation should be able to effectually reduce variable dependencies such that the resulting eigenvariables can be easily divided and the corresponding subpopulations can be efficiently evolved. Second, the transformation should be reversible [21]. As a counterpart of the forward transformation, the backward transformation transforms the offspring population generated in the eigenspace back into the original space such that it can be evaluated by the original fitness function.

The key of space transformation is to construct an eigen coordinate system. It involves two important factors: the computing method and the solution samples supporting the desirable eigen coordinate system. This study employs singular value decomposition as the computing method because it is an efficient tool to find the relationships hiding in data and has been widely used in various research fields, such as machine learning [51] and signal processing [52]. By conducting SVD on a set of solutions, an orthonormal basis of vectors can be obtained to form the eigen coordinate system. As for solution samples, it is not advisable to specially generate new solutions for SVD due to their extra evaluation cost. Actually, population individuals generated during the optimization process naturally contain useful distribution information, but are underutilized. It is convenient and meaningful to mine and exploit this kind of useful information with SVD.

Based on the above considerations, we suggest building the eigen coordinate system by conducting SVD on a set of high-quality solutions selected from the populations in recent generations. Concretely, EDC maintains a solution pool (SP) to preserve these solutions. In generation $t$, SP is constructed as follows:

$$SP = (H^{(t)}, H^{(t-1)}, \ldots, H^{(t-l+1)}) \tag{1}$$

where $H^{(t)}$ denotes the set of high-quality solutions selected in the $t$th generation and $l$ determines the size of SP. This implies that EDC stores the solutions selected from the recent $l$ generations into SP. SP can be viewed as an $n \times (l \cdot |H|)$ matrix with each column representing an $n$-dimensional complete solution, and there are totally $l \cdot |H|$ solutions in SP, where $|H|$ denotes the number of high-quality solutions selected in each generation. Generally, EDC takes the top $100 \cdot \tau\%$ solutions in a population as high-quality solutions, where $\tau$ is a truncation selection ratio varying in the range (0, 1).

Note that the average of the solutions in SP is generally nonzero, it needs to be removed to formalize the computation of SVD [21]. By subtracting this average, a zero-centered solution pool $SP_{zc}$ can be obtained and then decomposed as follows:

$$SP_{zc} = U\Sigma V^T \tag{2}$$

where $U$ is an $n \times n$ orthonormal matrix with its columns being left singular vectors. These vectors can be used as the orthonormal basis for establishing the eigenspace. In fact, the same set of vectors can also be obtained by performing eigenvalue decomposition on the covariance matrix estimated based on the solutions in SP [53]. SVD is more preferable because it is more robust and does not require the calculation of the covariance matrix.

Moreover, the proposed eigenspace construction method also has other three potential advantages. Firstly, it exploits the high-quality solutions selected from the recent populations, and thus provides a more reliable foundation for the computation of SVD. If all the population individuals are used, the computation of SVD might be disturbed too much by the randomness of inferior solutions. Secondly, compared with the solutions selected just from the current generation, SP contains more cumulative distribution information, and thus can reveal variable relationships better. Finally, the introduction of historical solutions in SP also alleviates the dependency of SVD on the population size. With more available solutions, SVD can be more robust and precise, especially in a high-dimensional space. This is of great significance to ensure the effectiveness of the subsequent eigenspace optimization process.

By virtue of the matrix $U$, the population $P$ in the original space can be transformed into the eigenspace as follows:

$$P' = U^{-1}P = U^T P \tag{3}$$

where $P'$ represents the transformed population in the eigenspace with each column being a solution corresponding to $n$ eigenvariables. Benefiting from this transformation, EDC carries out decomposition and optimization in the eigenspace rather than in the original space. After evolving each subpopulation of $P'$ with an EA, a new population can be obtained in the eigenspace. As the fitness function of the problem is defined only in the original space, it is necessary to transform the obtained new $P'$ back into the original space to evaluate the solutions therein. The backward transformation can be simply achieved by

$$P = UP' \tag{4}$$

### 3.2 Problem decomposition

After transformation, the original LSOP can be viewed as a high-dimensional problem with many weakly dependent variables in the eigenspace. Nevertheless, the problem is still hard to solve due to its huge solution space. This necessitates decomposing the problem into multiple low-dimensional subproblems such that the whole search space can be significantly reduced. Considering that the dependencies among the eigenvariables are weak in the eigenspace, it is reasonable to randomly divide these eigenvariables into multiple disjoint subgroups and accordingly partition the transformed population into several subpopulations.

To be more specific, EDC randomly divides an $n$-dimensional eigenvariable vector $x' = (x'_1, x'_2, \ldots, x'_n)$ into $m$

$s$-dimensional subcomponents $x'(1) = (x'_{i_1}, ..., x'_{i_s})$, $x'(2) = (x'_{i_{s+1}}, ..., x'_{i_{2s}})$, ..., $x'(m) = (x'_{i_{(m-1)s+1}}, ..., x'_{i_n})$, where $(i_1, ..., i_n)$ is a random permutation of the natural sequence $(1, ..., n)$ and $s$ is a predefined parameter satisfying $m = \lceil n/s \rceil$. According to the decomposition result, the transformed population $P'$ can be easily partitioned into $m$ subpopulations $P'(1), ..., P'(m)$.

The random decomposition strategy is conducted once at each generation by EDC. Besides the effect in reducing the search space, this strategy is also beneficial to improving the exploration ability of EDC. This mainly profits from its randomness, which helps to achieve different decomposition results at different generations and thus to enhance the diversity of subpopulations.

### 3.3 Subproblem optimization

Benefiting from the decomposition operation, a traditional EA can be employed to separately evolve these low-dimensional subpopulations. This study adopts the Gaussian EDA based on the general second order mixed moment (GSM-GEDA) [54] as optimizer. As a new variant of GEDA, GSM-GEDA inherits the basic framework of GEDA, but significantly enhances its efficiency by improving the estimation method for the Gaussian model.

Concretely, GSM-GEDA employs the following Gaussian model to describe the distribution of high-quality solutions:

$$G_{(\mu,C)}(x) = \frac{1}{(2\pi)^{n/2}(\det C)^{1/2}} e^{-\frac{1}{2}(x-\mu)^T C^{-1}(x-\mu)} \tag{5}$$

where $\mu$ and $C$ are the mean and covariance matrix of $x$ and determine the search center and scope of GEDA, respectively. It is worthy to mention that although GSM-GEDA is employed to tackle subproblems in the eigenspace in this study, the symbols of the original problem such as $x$ are used here for the convenience of description.

To obtain a promising search center in a generation $t$, GSM-GEDA first pre-estimates $\mu$ as the weighted average of the high-quality solutions selected from the current population:

$$\tilde{\mu}^{(t)} = \frac{\sum_{i=1}^{|H^{(t)}|} [\log(|H^{(t)}|+1) - \log(i)] H_{(i)}^{(t)}}{\sum_{i=1}^{|H^{(t)}|} [\log(|H^{(t)}|+1) - \log(i)]} \tag{6}$$

where $H_{(i)}^{(t)}$ denotes the $i$th best solution in the selected high-quality solution set $H^{(t)}$. From Eq. (6), it can be seen that the better a solution is, the larger its weight is. This operation tends to lead $\tilde{\mu}^{(t)}$ to a more promising solution region. To accelerate the search process, GSM-GEDA further shifts $\tilde{\mu}^{(t)}$ along its evolution direction as follows:

$$\tilde{\delta}^{(t)} = \tilde{\mu}^{(t)} - \hat{\mu}^{(t-1)} \tag{7}$$

$$\hat{\mu}^{(t)} = \begin{cases} \tilde{\mu}^{(t)} + \eta_f \tilde{\delta}^{(t)}, & f(\tilde{\mu}^{(t)} + \eta_f \tilde{\delta}^{(t)}) < f(\tilde{\mu}^{(t)}) < f(\hat{\mu}^{(t-1)}) \\ \tilde{\mu}^{(t)} - \eta_b \tilde{\delta}^{(t)}, & \max\{f(\tilde{\mu}^{(t)} - \eta_b \tilde{\delta}^{(t)}), f(\hat{\mu}^{(t-1)})\} < f(\tilde{\mu}^{(t)}) \\ \tilde{\mu}^{(t)}, & \text{otherwise} \end{cases} \tag{8}$$

where $\hat{\mu}^{(t)}$ is the final mean used in the $t$th generation, $\tilde{\delta}^{(t)}$ reflects the current evolution direction, and $\eta_f$ and $\eta_b$ are called the forward and backward shifting factor, respectively. Proper values for $\eta_f$ and $\eta_b$ could help GSM-GEDA finding a better search center and thus improve the search efficiency. The recommended values are $\eta_f = 2$ and

$\eta_b = 0.5$ [54]. It should be noted that the calculation of $\hat{\boldsymbol{\mu}}^{(t)}$ depends on $\hat{\boldsymbol{\mu}}^{(t-1)}$, and GSM-GEDA initializes $\hat{\boldsymbol{\mu}}^{(0)}$ as the average of the whole population in the first generation.

After determining the search center $\hat{\boldsymbol{\mu}}^{(t)}$, GSM-GEDA estimates the covariance matrix $C$ by taking $\hat{\boldsymbol{\mu}}^{(t)}$ as the reference mean:

$$\hat{C}^{(t)} = \frac{1}{|H^{(t)}|} \sum_{i=1}^{|H^{(t)}|} (H_i^{(t)} - \hat{\boldsymbol{\mu}}^{(t)})(H_i^{(t)} - \hat{\boldsymbol{\mu}}^{(t)})^{\mathrm{T}} \tag{9}$$

where $H_i^{(t)}$ is the $i$th solution in $H^{(t)}$. Based on $\hat{\boldsymbol{\mu}}^{(t)}$ and $\hat{C}^{(t)}$, a Gaussian probabilistic model $G^{(t)}$ could be built to sample new solutions for the next generation. The improved model estimation method in GSM-GEDA could effectively adjust its search scope and direction, and thus significantly enhances its optimization ability [54].

### 3.4 Procedure of EDC

By combining the space transformation technique, the problem decomposition strategy and the optimizer together, Algorithm 1 presents the detailed procedure of EDC, where four points deserve some attention. 1) EDC initializes $SP$ as empty in step 2. When updating $SP$ in step 6, EDC continuously stores the high-quality solutions selected in the first $l$ generations into $SP$. Once $SP$ becomes full, the oldest solutions in it will be replaced with the new ones. 2) EDC initializes the transformation matrix $U$ as an $n$-dimensional identity matrix $I$ in step 2 and updates $U$ every $l$ generations by conducting SVD in steps 7-9, where $\mathrm{mod}(t, l)$ denotes the remainder of $t$ divided by $l$. The motivation behind this operation is that only a tiny fraction of solutions in $SP$ change between two consecutive generations and accordingly the eigen coordinate system varies little. It is unnecessary to conduct SVD in each generation. Our preliminary experimental results showed that a proper value for $l$ could not only guarantee satisfactory performance, but also relieve the computation burden since it greatly reduces the execution frequency of SVD. 3) To build the Gaussian model for each subproblem, EDC first estimates the complete mean $\hat{\boldsymbol{\mu}}^{(t)}$ of all the variables in step 10, and then transforms $\hat{\boldsymbol{\mu}}^{(t)}$ into the eigenspace and partitions it into several subcomponents for different subproblems in steps 11-13. This is more convenient than separately estimating the mean for each subproblem. Moreover, EDC only transforms the solutions in $H^{(t)}$ rather than the ones in $P^{(t)}$ into the eigenspace in step 11 since the estimation of the covariance matrix is merely based on high-quality solutions. 4) EDC employs an elite strategy which maintains the best solution obtained so far to the next generation in step 21. So it only generates $p$-1 new subsolutions in step 17.

The complexity of EDC mainly comes from the SVD operation and the operations of GSM-GEDA. SVD can construct an $n$-dimensional eigenspace with a complexity of $O(n^3)$, and GSM-GEDA requires $O(s^3)$ computation quantity to evolve an $s$-dimensional subpopulation for a generation. Therefore, the total complexity of EDC in a generation is $O(n^3+ms^3) = O(n^3)$, which is the same as that of multivariate GEDAs. From the perspective of implementation, SVD is not executed in each generation but every $l$ generations. As indicated in the following experimental analysis, $l$ is generally set to a value far greater than 1, which would significantly reduce the computation burden of EDC. Moreover, it is notable that for many real-world LSOPs, the fitness evaluation of a solution is much more time consuming than algorithmic operations, and the efficiency of an algorithm is not so sensitive to its complexity. According to the above analysis, the complexity of EDC is definitely acceptable.

In comparison with existing DC-based methods, EDC can not only sharply shrink the search space of an LSOP, but also shows some distinctive characteristics that worth a few more discussion. Firstly, it significantly reduces the requirements on the decomposer and optimizer by employing a space transformation technique to convert the original problem into an eigenspace, where variable dependencies are greatly weakened. Secondly, EDC carries out a fine search in each low-dimensional eigensubspace with a novel variant of multivariate GEDA and conducts an explorative search in the whole solution space by randomly partitioning eigensubspaces in each generation. By this means, it achieves a balance between exploration and exploitation from a spatial perspective. Thirdly, EDC neither converts the original problem into a fixed eigenspace during the whole optimization process nor updates the eigenspace in each generation. Instead, it regularly updates the eigenspace when the pool of high-quality solutions is completely renewed in a certain number of generations. Profiting from this mechanism, EDC does not ask for an accurate enough eigenspace, which lowers its requirements on the space transformation technique and the population size. Finally, EDC concurrently and independently evolves all the subpopulations in one generation, which allows parallel computing to speed up the whole optimization process. Whereas CC only optimizes a single subproblem at a time with a context vector-based subsolution evaluation method.

---

**Algorithm 1**: Procedure of EDC

1. Initialize parameters, including population size $p$, selection ratio $\tau$, the size of solution pool $l$ and subproblem size $s$;
2. Set $t=1$, $SP=\emptyset$, $U=I$, and randomly generate the initial population $P^{(t)}$;
3. **while** the stopping criterion is not met **do**
4.     Evaluate $P^{(t)}$ and update the best solution $b^{(t)}$ obtained so far;
5.     Select the best $\lfloor \tau p \rfloor$ solutions from $P^{(t)}$ and store them into $H^{(t)}$;
6.     Update $SP$ with $H^{(t)}$ according to Eq. (1);
7.     **if** $mod(t, l)==0$    // update $U$ every $l$ generations
8.       Conduct SVD on $SP$ and update $U$ according to Eq. (2);
9.     **end if**
10.     Estimate $\hat{\mu}^{(t)}$ according to Eqs. (6)-(8);
11.     Transform $H^{(t)}$ and $\hat{\mu}^{(t)}$ into the eigenspace using Eq. (3) and obtain $H'^{(t)}$ and $\hat{\mu}'^{(t)}$;   // space transformation
12.     Generate a random decomposition $x' \to \{x'(1),...,x'(m)\}$ with $m = \lceil n/s \rceil$;
13.     Divide $H'^{(t)}$ and $\hat{\mu}'^{(t)}$ into $m$ subcomponents $H'^{(t)}(1),...,H'^{(t)}(m)$ and $\hat{\mu}'^{(t)}(1),...,\hat{\mu}'^{(t)}(m)$, respectively, according to the decomposition result;
14.     **for** $i=1$ **to** $m$   // evolve each subpopulation separately with GSM-GEDA
15.       Estimate covariance matrix $\hat{C}'^{(t)}(i)$ with $H'^{(t)}(i)$ and $\hat{\mu}'^{(t)}(i)$ using Eq. (9);
16.       Build a Gaussian model $G'^{(t)}(i)$ based on $\hat{\mu}'^{(t)}(i)$ and $\hat{C}'^{(t)}(i)$;
17.       Generate $p-1$ new subsolutions by sampling from $G'^{(t)}(i)$ and store them into $Q'^{(t)}(i)$;
18.     **end for**
19.     Generate a complete solution set $Q'^{(t)}$ by merging $Q'^{(t)}(1),...,Q'^{(t)}(m)$ together;
20.     Transform $Q'^{(t)}$ back into the original space using Eq. (4) and obtain $Q^{(t)}$;   // space transformation
21.     Set $P^{(t+1)} = Q^{(t)} \cup b^{(t)}$ and update $t = t+1$;
22. **end while**
23. **return** $b$.

---

## 4. Empirical Studies

To test the performance of EDC, this study conducted a series of experiments, where the first 14 benchmark functions (denoted as CEC'2005$_1$−CEC'2005$_{14}$) from the CEC'2005 test suite [23] and all the 20 ones (denoted as CEC'2010$_1$−CEC'2010$_{20}$) from the CEC'2010 large-scale test suite [6] were employed. CEC'2005$_1$−CEC'2005$_{14}$ can be set with different dimensions such as 100, 200 and 500. CEC'2010$_1$−CEC'2010$_{20}$ all have the same dimension of 1000, and are believed to be much harder to solve. Table 1 summaries the main characteristics of the two sets of test functions. More details of them can be found in [23] and [6].

In all the experiments, a maximum number of fitness evaluations (Max_FEs) was used as the termination condition of an algorithm. For functions CEC'2005$_1$−CEC'2005$_{14}$ with 100, 200 and 500 dimensions, Max_FEs was set to $1\times10^6$, $2\times10^6$ and $3\times10^6$, respectively; and for functions CEC'2010$_1$−CEC'2010$_{20}$, Max_FEs was set to $3\times10^6$. Unless otherwise mentioned, an algorithm was independently run 25 times on each function and was evaluated by the function error value (FEV), which is defined as the difference between the achieved best fitness value and the true optimum. Note that FEVs smaller than $10^{-8}$ are reported as zero in this study.

Table 1. Main characteristics of the test functions from the CEC'2005 test suite and the CEC'2010 large-scale test suite.

| CEC'2005 test suite | | | CEC'2010 larges-scale test suite | | |
| --- | --- | --- | --- | --- | --- |
| Function | Modality | Separability | Function | Modality | Separability |
| CEC'2005$_1$ | unimodal | separable | CEC'2010$_1$ | unimodal | separable |
| CEC'2005$_2$ | unimodal | nonseparable | CEC'2010$_2$ | multimodal | separable |
| CEC'2005$_3$ | unimodal | nonseparable | CEC'2010$_3$ | multimodal | separable |
| CEC'2005$_4$ | unimodal | nonseparable | CEC'2010$_4$ | unimodal | partially separable |
| CEC'2005$_5$ | unimodal | nonseparable | CEC'2010$_5$ | multimodal | partially separable |
| CEC'2005$_6$ | multimodal | nonseparable | CEC'2010$_6$ | multimodal | partially separable |
| CEC'2005$_7$ | multimodal | nonseparable | CEC'2010$_7$ | unimodal | partially separable |
| CEC'2005$_8$ | multimodal | nonseparable | CEC'2010$_8$ | multimodal | partially separable |
| CEC'2005$_9$ | multimodal | separable | CEC'2010$_9$ | unimodal | partially separable |
| CEC'2005$_{10}$ | multimodal | nonseparable | CEC'2010$_{10}$ | multimodal | partially separable |
| CEC'2005$_{11}$ | multimodal | nonseparable | CEC'2010$_{11}$ | multimodal | partially separable |
| CEC'2005$_{12}$ | multimodal | nonseparable | CEC'2010$_{12}$ | unimodal | partially separable |
| CEC'2005$_{13}$ | multimodal | nonseparable | CEC'2010$_{13}$ | multimodal | partially separable |
| CEC'2005$_{14}$ | multimodal | nonseparable | CEC'2010$_{14}$ | unimodal | partially separable |
| | | | CEC'2010$_{15}$ | multimodal | partially separable |
| | | | CEC'2010$_{16}$ | multimodal | partially separable |
| | | | CEC'2010$_{17}$ | unimodal | partially separable |
| | | | CEC'2010$_{18}$ | multimodal | partially separable |
| | | | CEC'2010$_{19}$ | unimodal | nonseparable |
| | | | CEC'2010$_{20}$ | multimodal | nonseparable |

**4.1 Sensitivity to parameters**

There are four parameters in EDC, including the size of solution pool $l$, population size $p$, subproblem size $s$ and selection ratio $\tau$. In all the experiments, $\tau$ was conventionally set to 0.5. This subsection mainly presents the sensitive of EDC with respect to the other three parameters, i.e, $l$, $p$ and $s$. Among them, $l$ determines the number of historical high-quality solutions used for computing the eigenspace, and thus affects the quality of the established eigenspace. $p$ determines the exploration ability of the optimizer in EDC, and for a given selection ratio $\tau$, it also determines the exploitation degree on high-quality solutions produced in each generation. As for $s$, it determines the solving difficulty of each subproblem by varying its size and affects the subproblem optimization process. In order to intensively investigate the influence of the three parameters, two separate experiments were implemented on the CEC'2005 test suite that covers different types of functions. In the first experiment, we varied the value of $l$ while keeping $s$ and $p$ at fixed values to study the performance variation of EDC with respect to $l$. In the second experiment, we investigated the sensitivity of EDC to both $s$ and $p$ while keeping $l$ at a fixed value. For simplicity, only the experimental results on two representative functions, i.e. CEC'2005$_4$ and CEC'2005$_9$, were presented.

*1) Sensitivity to l.* We tested the performance variation of EDC when setting $l$ to different values in {2,10,20,30,40} and fixing $p$ and $s$ at the following default values: $p = 1000$ and $s = 30$. Fig. 2 shows the evolution curves of the average FEVs achieved by EDC, where the two functions CEC'2005$_4$ and CEC'2005$_9$ with 100 dimensions ($D = 100$) are taken as examples. It can be seen that EDC obtains desirable final results when $l$ varies from 10 to 30, which means that it is rather robust to $l$. The evolution trends of FEVs further indicate that $l$ has great influence on the convergence ability of EDC. A small value for $l$ could accelerate the convergence process, but may cause premature convergence. On the contrary, a large value for $l$ is helpful to improve the exploration ability of EDC, but tends to slow down the convergence speed. Fig. 3 presents the results obtained by EDC on the two functions with $D = 200$. Its comparison with Fig. 2 indicates that EDC has similar robustness to $l$ when the function dimension increases. This observation also reveals that EDC is rather scalable to the problem dimension. To balance the performance of EDC in different aspects, this study suggests setting $l$ to a value within [10, 30].

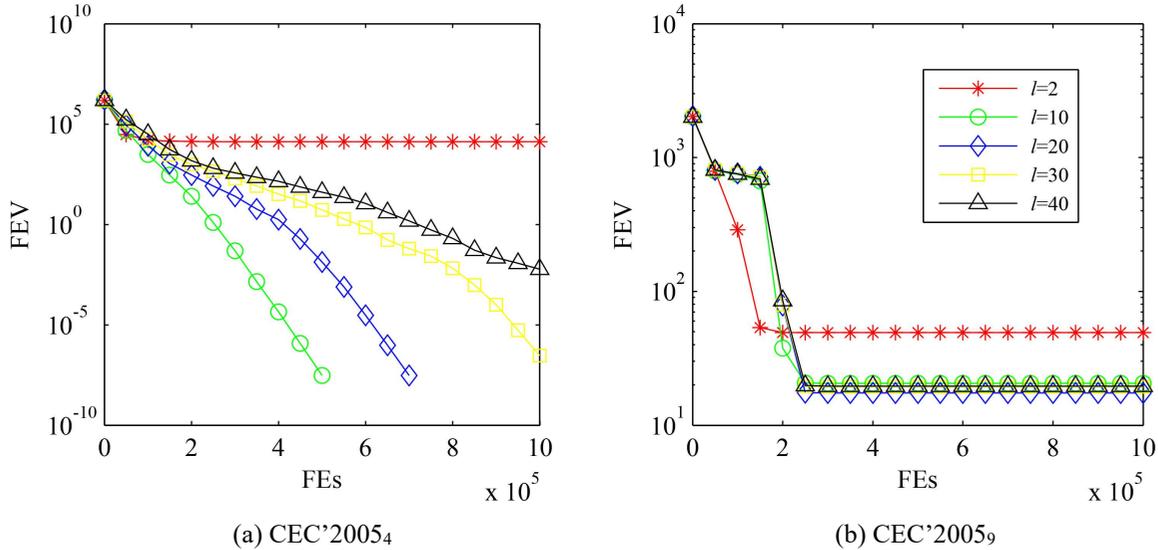

Figure 2. Performance of EDC with different $l$ values on functions with $D = 100$.

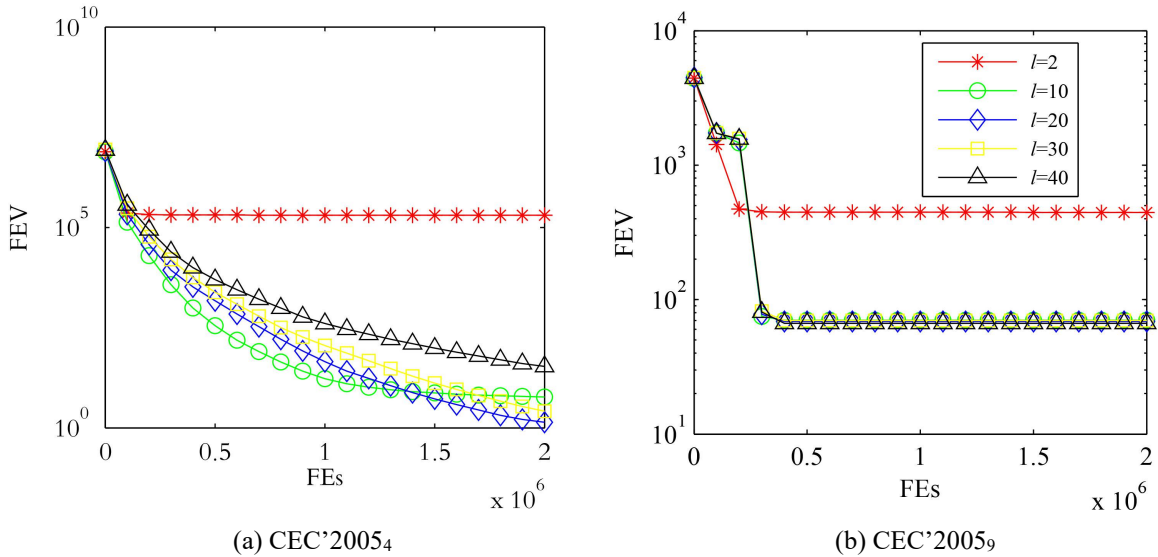

Figure 3. Performance of EDC with different $l$ values on functions with $D = 200$.

*2) Sensitivity to p and s.* To investigate the influence of $p$ and $s$, we evaluated the performance of EDC with different combinations of $p$ and $s$ while keeping $l$ at a fixed value of 20. The optional values for $p$ and $s$ in this experiment were $p \in \{500, 1000, 1500, 2000\}$ and $s \in \{10, 20, 30, 40, 50\}$. Figs. 4 and 5 present the results obtained by EDC on the functions CEC'2005$_4$ and CEC'2005$_9$ with different dimensions. The comparison between these two figures indicates that the combination of $p$ and $s$ shows similar influence on the performance of EDC when the problem dimension varies. Concretely, for the unimodal nonseparable function CEC'2005$_4$, EDC can get desirable solutions with different $s$ values as long as a matching $p$ value is provided. Its performance deteriorates a little only when a large $s$ is matched with a small $p$, and vice verse. The underlying reason consists in that a small population cannot adequately explore the solution space of a large-scale subproblem; inversely, for a small scale subproblem, an overlarge population greatly reduces iteration times of the optimizer, which is also detrimental to the performance of EDC. When it comes to the multimodal separable function CEC'2005$_9$, EDC demonstrates distinctively different performance. Benefiting by the separability of this function, EDC achieves almost the same solution when $s$ changes from 10 to 50, while it generally requires a large population to cope with the multimodality of this function. Based on these observations, this study recommends to set $p$ within [1000, 1500] and $s$ within [30, 40].

Through the above two experiments, it can be concluded that EDC is rather robust to its parameters and has fine scalability to problem dimension. For the experiments described below, the three parameters were set as $l = 20$, $p = 1000$ and $s = 30$.

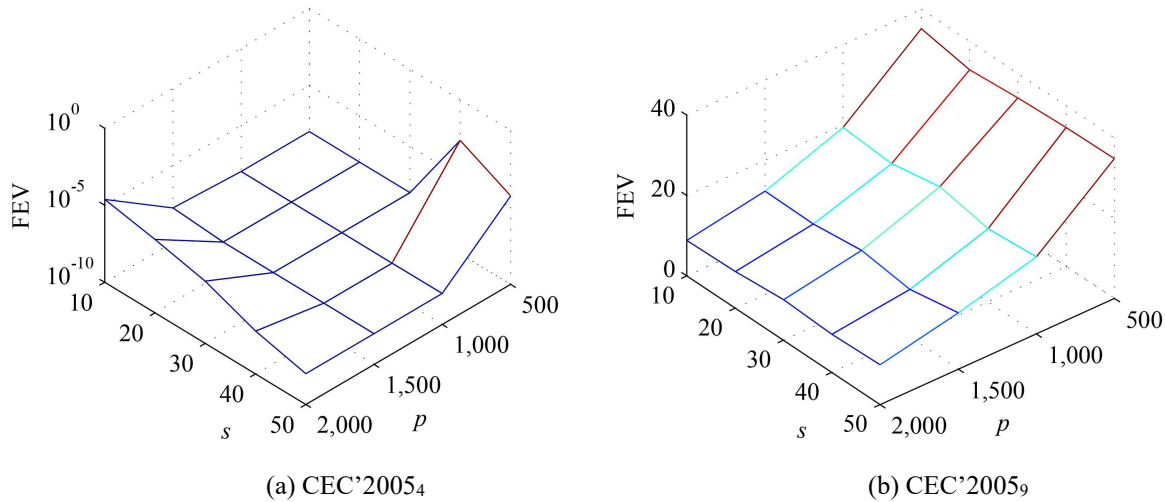

(a) CEC'2005$_4$          (b) CEC'2005$_9$

Figure 4. Performance of EDC with different combinations of $s$ and $p$ on functions with $D = 100$.

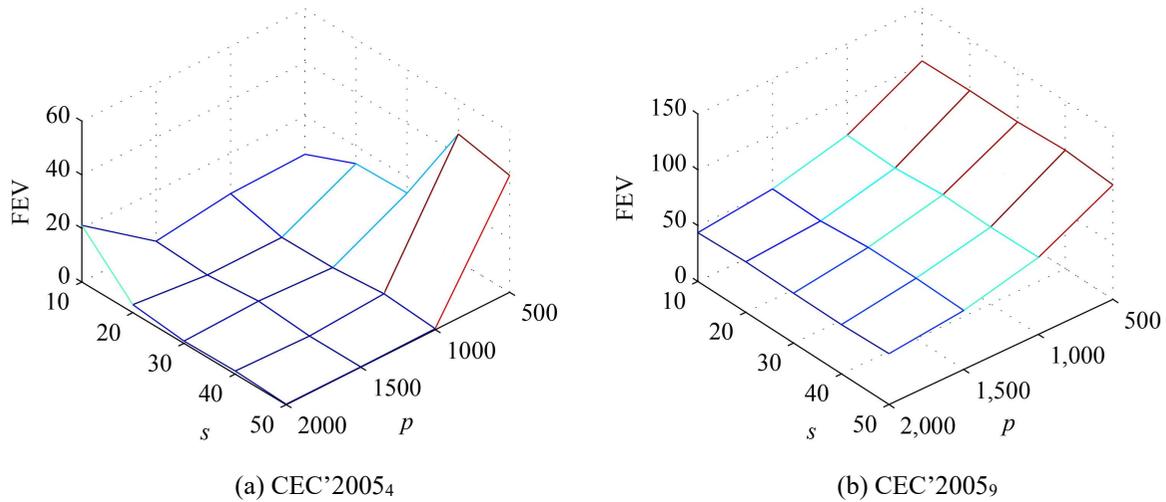

(a) CEC'2005$_4$          (b) CEC'2005$_9$

Figure 5. Performance of EDC with different combinations of $s$ and $p$ on functions with $D = 200$.

## 4.2 Experiments on CEC'2005 test suite

In this subsection, the performance of EDC was assessed on 14 test functions from CEC'2005 test suite. As EDC employs GSM-GEDA as the optimizer, three EDA-based algorithms for large-scale optimization, including EDA-MCC [43], EDA-MCC-MI [44] and RP-EDA [49], were selected for comparison. And the canonical GSM-GEDA was also included in this experiment to serve as a baseline. EDA-MCC, EDA-MCC-MI and RP-EDA have all been introduced in Section 2. To ensure the fairness of the comparison, we directly employed the source codes of EDA-MCC and EDA-MCC-MI provided by Xu et al. [44] and the source code of RP-EDA provided by Kabán et al. [49], and set their parameters to default values. As for GSM-GEDA, it was originally developed for solving low- and medium-dimensional problems. To make it perform better on high-dimensional problems, we set its population size to 2000 in this experiment. Its other parameters were designated as the suggestions in [54].

Tables 2-4 summarize the mean values and standard deviations of the FEVs obtained by EDA-MCC, EDA-MCC-MI, RP-EDA, GSM-GEDA and EDC on the 14 functions with 100, 200 and 500 dimensions, respectively. Wilcoxon's rank sum test at a 0.05 significance level was conducted between EDC and each of its competitors to measure their performance difference. The symbols "−", "≈" and "+" in Tables 2-4 indicate that the performance of the corresponding competitor is worse than, similar to and better than that of EDC, respectively. In addition, the best result obtained for each function is highlighted in bold.

Table 2. The mean values and standard deviations (mean±standard deviation) of the FEVs obtained by EDA-MCC, EDA-MCC-MI, RP-EDA, GSM-GEDA and EDC over 25 independent runs on 14 test functions from CEC'2005 test suite with $D = 100$. Statistical results are obtained based on Wilcoxon's rank sum test at a significance level of 0.05.

| Function | EDA-MCC | EDA-MCC-MI | RP-EDA | GSM-GEDA | EDC |
|---|---|---|---|---|---|
| CEC'2005$_1$ | **0.00E+00±0.00E+00** ≈ | **0.00E+00±0.00E+00** ≈ | **0.00E+00±0.00E+00** ≈ | **0.00E+00±0.00E+00** ≈ | **0.00E+00±0.00E+00** |
| CEC'2005$_2$ | 7.12E+04±3.64E+03 − | 7.14E+04±6.77E+03 − | 3.16E+04±7.20E+03 − | **0.00E+00±0.00E+00** ≈ | **0.00E+00±0.00E+00** |
| CEC'2005$_3$ | 6.22E+07±1.93E+07 − | 6.11E+07±2.02E+07 − | 7.41E+06±1.84E+06 − | 2.88E+04±2.29E+04 − | **1.99E+01±2.59E+01** |
| CEC'2005$_4$ | 9.15E+04±6.64E+03 − | 9.04E+04±7.36E+03 − | 4.70E+04±8.46E+03 − | 4.81E+03±2.62E+03 − | **0.00E+00±0.00E+00** |
| CEC'2005$_5$ | 1.23E+04±4.78E+02 − | 1.24E+04±4.65E+02 − | 8.43E+03±7.53E+02 − | 4.26E+02±2.96E+02 − | **2.17E−04±6.91E−04** |
| CEC'2005$_6$ | 1.01E+05±7.15E+04 − | 1.20E+05±1.28E+05 − | 9.60E+01±9.76E−01 − | 9.36E+01±2.08E+00 − | **4.58E+01±4.66E−01** |
| CEC'2005$_7$ | 1.44E+01±8.01E+00 − | 1.06E+01±5.94E+00 − | 8.03E−03±7.60E−03 − | **0.00E+00±0.00E+00** ≈ | **0.00E+00±0.00E+00** |
| CEC'2005$_8$ | **2.13E+01±2.02E−02** ≈ | **2.13E+01±1.54E−02** ≈ | **2.13E+01±2.37E−02** ≈ | **2.13E+01±2.87E−02** ≈ | **2.13E+01±2.63E−02** |
| CEC'2005$_9$ | **1.46E+01±2.22E+00** + | 1.60E+01±3.20E+00 + | 7.29E+02±1.58E+01 − | 3.89E+01±5.03E+00 − | 1.95E+01±5.40E+00 |
| CEC'2005$_{10}$ | 2.07E+01±4.09E+00 − | 2.14E+01±4.41E+00 − | 7.42E+02±9.54E+00 − | 4.94E+01±1.26E+01 − | **1.36E+01±4.85E+00** |
| CEC'2005$_{11}$ | 2.80E+01±3.63E+00 − | 2.84E+01±3.29E+00 − | 1.91E+01±2.74E+00 − | 2.03E+01±4.07E+00 − | **1.84E+00±2.09E+00** |
| CEC'2005$_{12}$ | 5.06E+05±1.23E+05 − | 5.11E+05±8.47E+04 − | 1.39E+05±8.10E+04 − | 4.35E+04±2.41E+04 − | **1.69E+04±1.69E+04** |
| CEC'2005$_{13}$ | 1.32E+01±9.86E−01 − | 1.31E+01±1.05E+00 − | 6.50E+01±1.69E+00 − | 1.20E+01±1.25E+00 − | **1.06E+01±9.14E−01** |
| CEC'2005$_{14}$ | 4.73E+01±2.65E−01 − | 4.73E+01±3.49E−01 − | 4.70E+01±2.13E−01 − | **4.57E+01±3.98E−01** ≈ | **4.57E+01±2.65E−01** |
| −/≈/+ Nos. | 11/2/1 | 11/2/1 | 12/2/0 | 9/5/0 | − |

"−", "≈" and "+" denote that the performance of the corresponding algorithm is worse than, similar to and better than that of EDC, respectively.

Table 3. The mean values and standard deviations (mean±standard deviation) of the FEVs obtained by EDA-MCC, EDA-MCC-MI, RP-EDA, GSM-GEDA and EDC over 25 independent runs on 14 test functions from CEC'2005 test suite with $D = 200$. Statistical results are obtained based on Wilcoxon's rank sum test at a significance level of 0.05.

| Function | EDA-MCC | EDA-MCC-MI | RP-EDA | GSM-GEDA | EDC |
|---|---|---|---|---|---|
| CEC'2005$_1$ | **0.00E+00±0.00E+00** ≈ | **0.00E+00±0.00E+00** ≈ | **0.00E+00±0.00E+00** ≈ | 2.70E+03±6.15E+02 − | **0.00E+00±0.00E+00** |
| CEC'2005$_2$ | 2.94E+05±2.69E+04 − | 3.19E+05±3.01E+04 − | 3.66E+04±5.09E+03 − | 3.62E+04±5.90E+03 − | **0.00E+00±0.00E+00** |
| CEC'2005$_3$ | 1.64E+08±3.02E+07 − | 1.71E+08±2.80E+07 − | 2.47E+07±5.12E+06 − | 1.91E+07±6.49E+06 − | **1.79E+02±1.40E+02** |
| CEC'2005$_4$ | 4.06E+05±6.48E+04 − | 4.02E+05±4.37E+04 − | 9.40E+04±1.67E+04 − | 1.78E+05±2.64E+04 − | **2.18E+00±7.51E+00** |
| CEC'2005$_5$ | 3.00E+04±1.94E+03 − | 2.87E+04±1.70E+03 − | 2.37E+04±1.71E+03 − | 1.85E+04±2.26E+03 − | **3.18E+02±2.80E+02** |
| CEC'2005$_6$ | 3.59E+05±2.70E+05 − | 2.81E+05±2.72E+05 − | 4.52E+02±1.26E+03 − | 7.56E+06±3.33E+06 − | **1.29E+02±2.63E+01** |
| CEC'2005$_7$ | 7.82E+01±2.85E+01 − | 7.61E+01±3.49E+01 − | 7.90E+02±1.87E+02 − | 2.06E+02±2.12E+02 − | **4.63E−03±2.31E−02** |
| CEC'2005$_8$ | **2.14E+01±1.86E−02** ≈ | **2.14E+01±1.35E−02** ≈ | **2.14E+01±1.75E−02** ≈ | **2.14E+01±1.75E−02** ≈ | **2.14E+01±1.68E−02** |
| CEC'2005$_9$ | 5.70E+01±6.04E+00 + | **5.68E+01±7.17E+00** + | 9.81E+01±1.23E+01 − | 4.85E+02±3.79E+01 − | 6.89E+01±9.89E+00 |
| CEC'2005$_{10}$ | 7.10E+01±7.48E+00 − | 6.68E+01±7.71E+00 − | 9.67E+01±1.18E+01 − | 6.05E+02±4.00E+01 − | **5.32E+01±8.35E+00** |
| CEC'2005$_{11}$ | 6.11E+01±6.02E+00 − | 6.00E+01±5.65E+00 − | 5.50E+01±5.88E+00 − | 1.32E+02±7.11E+00 − | **6.55E+00±5.06E+00** |
| CEC'2005$_{12}$ | 2.33E+06±5.51E+05 − | 2.34E+06±5.48E+05 − | 1.09E+06±5.04E+05 − | 2.30E+06±5.26E+05 − | **8.05E+04±5.36E+04** |
| CEC'2005$_{13}$ | 2.28E+01±1.17E+00 − | 2.26E+01±1.29E+00 − | 2.18E+01±1.34E+00 ≈ | 3.62E+01±3.44E+00 − | **2.11E+01±1.47E+00** |
| CEC'2005$_{14}$ | 9.65E+01±3.80E−01 − | 9.65E+01±4.80E−01 − | 9.57E+01±4.25E−01 − | **9.47E+01±5.68E−01** ≈ | 9.48E+01±3.69E−01 |
| −/≈/+ Nos. | 11/2/1 | 11/2/1 | 11/3/0 | 12/2/0 | − |

"−", "≈" and "+" denote that the performance of the corresponding algorithm is worse than, similar to and better than that of EDC, respectively.

Table 4. The mean values and standard deviations (mean±standard deviation) of the FEVs obtained by EDA-MCC, EDA-MCC-MI, RP-EDA, GSM-GEDA and EDC over 25 independent runs on 14 test functions from CEC'2005 test suite with $D = 500$. Statistical results are obtained based on Wilcoxon's rank sum test at a significance level of 0.05.

| Function | EDA-MCC | EDA-MCC-MI | RP-EDA | GSM-GEDA | EDC |
|---|---|---|---|---|---|
| CEC'2005$_1$ | **0.00E+00±0.00E+00** ≈ | **0.00E+00±0.00E+00** ≈ | **0.00E+00±0.00E+00** ≈ | 2.40E+05±2.13E+04 − | **0.00E+00±0.00E+00** |
| CEC'2005$_2$ | 1.46E+06±1.28E+05 − | 1.49E+06±1.38E+05 − | 3.87E+05±2.71E+04 − | 7.50E+05±5.97E+04 − | **0.00E+00±0.00E+00** |
| CEC'2005$_3$ | 4.78E+08±4.32E+07 − | 4.38E+08±5.78E+07 − | 1.18E+08±2.13E+07 − | 1.40E+09±2.24E+08 − | **4.96E+04±1.24E+04** |
| CEC'2005$_4$ | 1.98E+06±1.87E+05 − | 1.97E+06±2.18E+05 − | 9.82E+05±5.44E+04 − | 1.45E+06±1.55E+05 − | **1.19E+04±4.27E+03** |
| CEC'2005$_5$ | 6.46E+04±2.23E+03 − | 6.04E+04±2.24E+03 − | 5.27E+04±2.21E+03 − | 9.56E+04±4.29E+03 − | **6.86E+03±9.68E+02** |
| CEC'2005$_6$ | 4.50E+06±2.33E+06 − | 9.35E+05±6.88E+05 − | 2.79E+03±3.72E+03 − | 2.11E+10±3.41E+09 − | **4.27E+02±3.52E+01** |
| CEC'2005$_7$ | 3.14E+02±8.83E+01 − | 1.96E+02±6.32E+01 − | 2.99E+03±1.23E+02 − | 3.79E+04±3.63E+03 − | **0.00E+00±0.00E+00** |
| CEC'2005$_8$ | **2.15E+01±9.34E−03** ≈ | **2.15E+01±1.13E−02** ≈ | **2.15E+01±6.72E−03** ≈ | **2.15E+01±6.24E−03** ≈ | 2.15E+01±7.83E−03 |
| CEC'2005$_9$ | 2.20E+02±1.77E+01 + | **2.00E+02±1.40E+01** + | 3.35E+02±2.11E+01 − | 3.27E+03±2.71E+02 − | 2.44E+02±2.38E+01 |
| CEC'2005$_{10}$ | 2.59E+02±1.93E+01 − | 2.31E+02±1.33E+01 − | 3.87E+02±3.65E+01 − | 4.04E+03±1.65E+02 − | **2.09E+02±1.95E+01** |
| CEC'2005$_{11}$ | 1.74E+02±9.58E+00 − | 1.62E+02±8.55E+00 − | 1.71E+02±5.03E+00 − | 5.75E+02±1.23E+01 − | **4.14E+01±1.05E+01** |
| CEC'2005$_{12}$ | 1.61E+07±1.88E+06 − | 1.45E+07±2.16E+06 − | 1.42E+07±2.52E+06 − | 1.39E+08±1.30E+07 − | **3.48E+05±1.52E+05** |
| CEC'2005$_{13}$ | 5.53E+01±2.22E+00 − | 5.53E+01±2.05E+00 − | 5.60E+01±2.29E+00 − | 1.88E+02±8.63E+00 − | **5.38E+01±2.28E+00** |
| CEC'2005$_{14}$ | 2.45E+02±4.49E−01 − | 2.45E+02±3.77E−01 − | **2.43E+02±4.67E−01** + | **2.43E+02±3.83E−01** ≈ | 2.44E+02±3.28E−01 |
| −/≈/+ Nos. | 11/2/1 | 11/2/1 | 11/2/1 | 12/2/0 | − |

"−", "≈" and "+" denote that the performance of the corresponding algorithm is worse than, similar to and better than that of EDC, respectively.

According to the experimental results in Tables 2-4, the following observations can be derived:

1) EDC achieves the overall best performance on this set of functions. Compared with EDA-MCC and EDA-MCC-MI, EDC could always obtain better results on 11 out of 14 functions and is only defeated by them on one function (CEC'2005$_9$), regardless of the problem dimension. The failure of EDC on CEC'2005$_9$ can be mainly attributed to the fact that CEC'2005$_9$ is separable, profiting from which, EDA-MCC and EDA-MCC-MI can properly decompose this function into multiple subproblems in the original space without any sophisticated operations, and further efficiently address it in the resulting subspaces. In contrast, by a space transformation technique, EDC sacrifices a little decomposition accuracy on separable problems to earn its strong robustness on nonseparable ones. Nevertheless, the final results obtained by EDC on CEC'2005$_9$ is still in the same order of magnitude with the corresponding ones obtained by EDA-MCC and EDA-MCC-MI. This indicates that EDC also takes effect on separable problems. With respect to RP-EDA and GSM-GEDA, EDC performs no worse than them on all the functions, except being surpassed by RP-EDA on CEC'2005$_{14}$ with $D = 500$. It's necessary to mention that RP-EDA was reported to perform well especially on multimodal functions [49]. The superiority of EDC over RP-EDA on the vast majority of multimodal functions demonstrates that EDC has good exploration ability in the high-dimensional space. What's more, EDC is also able to attain high-efficiency performance on unimodal functions. As can be seen in Tables 2-4, EDC consistently provides the best results for the five unimodal functions (CEC'2005$_1$-CEC'2005$_5$) with $D = 100$, 200 and 500, and for CEC'2005$_2$-CEC'2005$_5$, its results are better than those of its competitors by several orders of magnitude.

2) The eigenspace divide-and-conquer framework are very effective. EDC scales up GSM-GEDA by virtue of a space transformation technique and a decomposition strategy. The experimental results in Tables 2-4 show that EDC performs no worse than GSM-GEDA on all the test functions and its superiority becomes larger with the growth of problem dimension, which verifies the effectiveness and efficiency of the proposed eigenspace divide-and-conquer framework. EDA-MCC and EDA-MCC-MI also adopt a DC strategy, but conduct decomposition and optimization in the original solution space. Compared with them, EDC is able to achieve better performance by transforming the DC process into the eigenspace. Although the decomposer and optimizer in EDC are not exactly identical with those in EDA-MCC and EDA-MCC-MI, the significant superiority of EDC might still verify the efficiency of its space transformation technique to a certain extent. We will further investigate the performance of the space transformation technique in Section 4.4.

3) EDC is very robust and has good scalability to problem dimension. From Tables 2-4, it can be obviously seen that

EDC can stably find high-quality solutions for most functions when their dimensions vary from 100 to 500. The excellent performance of EDC mainly benefits from its eigenspace divide-and-conquer framework, which endows it strong adaptability to different problem characteristics and dimensions.

### 4.3 Comparison on CEC'2010 large-scale test suite

To further evaluate the performance of EDC on higher-dimensional problems, we tested it on the 20 1000-dimensional test functions from the CEC'2010 large-scale test suite [6] and compared it with seven state-of-the-art algorithms, including MA-SW-Chains [11], MOS [12], DSPLSO [47], Rm-ES [46], RP-EDA [49], MMO-CC [37] and NPDC [45]. Among these competitors, the first five algorithms are representatives of non-DC-based algorithms, and the latter two algorithms are efficient DC-based algorithms, they have all been briefly introduced in the literature review section.

Table 5. Experimental results obtained by MA-SW-Chains, MOS, DSPLSO, Rm-ES, RP-EDA, MMO-CC, NPDC and EDC over 25 independent runs on the 20 1000-dimensional functions from the CEC'2010 large-scale test suite. Statistical results are obtained based on Cohen'$d$ effect size and Friedman test.

| Function | | MA-SW-Chains | MOS | DSPLSO | Rm-ES | RP-EDA | MMO-CC | NPDC | EDC |
|---|---|---|---|---|---|---|---|---|---|
| CEC'2010$_1$ | Mean | **0.00E+00** + | **0.00E+00** + | **0.00E+00** + | 4.81E+06 − | 2.74E+08 − | **0.00E+00** + | **0.00E+00** + | 3.18E+06 |
| | Std | **0.00E+00** | **0.00E+00** | **0.00E+00** | 3.38E+05 | 9.57E+06 | **0.00E+00** | **0.00E+00** | 2.96E+05 |
| CEC'2010$_2$ | Mean | 8.10E+02 − | **1.97E+02** + | 4.45E+02 + | 4.89E+02 + | 9.03E+02 − | 1.43E+03 − | 8.38E+03 − | 5.62E+02 |
| | Std | 5.88E+01 | **1.59E+01** | 1.65E+01 | 2.37E+01 | 7.62E+01 | 8.43E+01 | 3.69E+02 | 4.50E+01 |
| CEC'2010$_3$ | Mean | **0.00E+00** ≈ | 1.12E+00 − | **0.00E+00** ≈ | 5.09E−04 − | **0.00E+00** ≈ | **0.00E+00** ≈ | 1.99E+01 − | **0.00E+00** |
| | Std | **0.00E+00** | 1.00E+00 | **0.00E+00** | 2.50E−04 | **0.00E+00** | **0.00E+00** | 1.29E−02 | **0.00E+00** |
| CEC'2010$_4$ | Mean | 3.53E+11 − | 1.91E+10 − | 4.30E+11 − | 4.03E+11 − | 1.03E+12 − | **7.64E+06** + | 1.60E+10 − | 2.95E+07 |
| | Std | 3.12E+10 | 8.08E+09 | 8.31E+10 | 3.13E+10 | 2.61E+11 | **1.31E+06** | 9.79E+09 | 2.49E+06 |
| CEC'2010$_5$ | Mean | 1.68E+08 − | 6.81E+08 − | 6.30E+06 − | 7.57E+07 − | 1.20E+07 − | 3.34E+08 − | 5.71E+08 − | **3.63E+06** |
| | Std | 1.04E+08 | 1.42E+08 | 1.76E+06 | 1.75E+07 | 3.03E+06 | 1.54E+08 | 1.54E+08 | **1.62E+06** |
| CEC'2010$_6$ | Mean | 8.14E+04 − | 2.00E+07 − | **0.00E+00** + | 2.16E+01 − | 2.13E+01 − | 5.77E−01 + | 1.98E+07 − | 2.09E+01 |
| | Std | 2.84E+05 | 5.67E+04 | **0.00E+00** | 1.41E−02 | 3.40E−02 | 1.32E+00 | 7.16E+04 | 3.95E−02 |
| CEC'2010$_7$ | Mean | 1.03E+02 + | **0.00E+00** + | 4.76E+02 + | 1.89E+07 − | 4.39E+06 − | 2.41E+10 − | **0.00E+00** + | 2.05E+04 |
| | Std | 8.70E+01 | **0.00E+00** | 1.31E+02 | 5.25E+06 | 1.96E+06 | 6.26E+09 | **0.00E+00** | 1.26E+04 |
| CEC'2010$_8$ | Mean | 1.41E+07 − | 1.12E+06 + | 3.11E+07 − | 8.21E+06 − | 1.56E+08 − | 2.63E+08 − | 1.20E+06 + | 1.97E+06 |
| | Std | 3.68E+07 | 1.79E+06 | 9.36E+04 | 4.81E+05 | 2.40E+08 | 5.29E+08 | 1.87E+06 | 1.25E+05 |
| CEC'2010$_9$ | Mean | 1.41E+07 − | 8.78E+06 − | 4.59E+07 − | 6.74E+06 − | 3.06E+08 − | **8.99E+01** + | 4.48E+06 − | 3.48E+06 |
| | Std | 1.15E+06 | 1.01E+06 | 3.04E+06 | 4.54E+05 | 1.76E+07 | **4.64E+01** | 5.12E+05 | 2.64E+05 |
| CEC'2010$_{10}$ | Mean | 2.07E+03 − | 7.86E+03 − | 7.99E+03 − | **4.92E+02** + | 9.57E+02 − | 1.63E+03 − | 1.23E+04 − | 6.01E+02 |
| | Std | 1.44E+02 | 2.43E+02 | 1.28E+02 | **1.53E+01** | 5.21E+01 | 9.10E+01 | 4.11E+02 | 3.31E+01 |
| CEC'2010$_{11}$ | Mean | 3.80E+01 − | 1.99E+02 − | **0.00E+00** + | 1.36E+02 − | 4.44E+01 − | 2.99E+00 + | 2.19E+02 − | 2.09E+01 |
| | Std | 7.35E+00 | 4.52E−01 | **0.00E+00** | 2.21E+01 | 6.63E+00 | 3.97E+00 | 2.49E−01 | 5.57E−02 |
| CEC'2010$_{12}$ | Mean | 3.62E−06 − | **0.00E+00** ≈ | 9.52E+04 − | **0.00E+00** ≈ | 5.73E+04 − | **0.00E+00** ≈ | **0.00E+00** ≈ | **0.00E+00** |
| | Std | 5.92E−07 | **0.00E+00** | 6.69E+03 | **0.00E+00** | 3.74E+03 | **0.00E+00** | **0.00E+00** | **0.00E+00** |
| CEC'2010$_{13}$ | Mean | 1.25E+03 − | 1.36E+03 − | 5.48E+02 + | 5.15E+05 − | 1.03E+06 − | 3.05E+04 − | **1.40E+01** + | 6.18E+02 |
| | Std | 5.72E+02 | 9.37E+02 | 1.69E+02 | 1.65E+05 | 4.50E+04 | 9.43E+04 | **1.38E+01** | 3.48E+02 |
| CEC'2010$_{14}$ | Mean | 3.11E+07 − | 1.82E+07 − | 1.60E+08 − | 6.24E+06 − | 3.90E+08 − | **0.00E+00** + | 1.79E+07 − | 3.50E+06 |
| | Std | 1.93E+06 | 1.18E+06 | 8.50E+06 | 9.43E+05 | 1.43E+07 | **0.00E+00** | 1.35E+06 | 3.21E+05 |
| CEC'2010$_{15}$ | Mean | 2.74E+03 − | 1.54E+04 − | 9.91E+03 − | **4.95E+02** + | 9.26E+02 − | 2.05E+03 − | 1.55E+04 − | 6.05E+02 |
| | Std | 1.22E+02 | 5.36E+02 | 6.70E+01 | **3.12E+01** | 3.90E+01 | 9.39E+01 | 4.54E+02 | 3.99E+01 |
| CEC'2010$_{16}$ | Mean | 9.98E+01 − | 3.97E+02 − | **0.00E+00** + | 1.42E+02 − | 9.40E+01 − | 8.87E+00 + | 3.98E+02 − | 1.48E+01 |
| | Std | 1.40E+01 | 2.10E−01 | **0.00E+00** | 3.82E+01 | 1.32E+01 | 9.25E+00 | 3.94E−01 | 7.33E+00 |
| CEC'2010$_{17}$ | Mean | 1.24E+00 − | 4.66E−05 + | 6.84E+05 − | **0.00E+00** + | 3.01E+05 − | **0.00E+00** + | 1.24E−06 + | 2.25E−02 |
| | Std | 1.25E−01 | 6.24E−06 | 3.63E+04 | **0.00E+00** | 1.38E+04 | **0.00E+00** | 7.56E−07 | 7.60E−02 |
| CEC'2010$_{18}$ | Mean | 1.30E+03 ≈ | 3.91E+03 − | 1.35E+03 − | 8.79E+02 + | 3.74E+04 − | 3.37E+04 − | **3.67E+02** + | 1.22E+03 |
| | Std | 4.36E+02 | 2.18E+03 | 3.87E+02 | 9.58E+02 | 9.71E+03 | 2.75E+04 | **2.20E+02** | 4.23E+02 |
| CEC'2010$_{19}$ | Mean | 2.85E+05 − | 3.41E+04 − | 8.20E+06 − | 1.36E+05 − | 1.82E+06 − | 1.54E+07 − | 1.07E+04 − | **5.27E+00** |
| | Std | 1.78E+04 | 2.63E+03 | 4.69E+05 | 3.69E+04 | 8.97E+04 | 1.59E+06 | 1.25E+03 | **3.08E+00** |
| CEC'2010$_{20}$ | Mean | 1.07E+03 − | 8.31E+02 − | 1.06E+03 − | 7.51E+02 + | 1.08E+03 − | 1.10E+03 − | **7.97E−01** + | 9.34E+02 |
| | Std | 7.29E+01 − | 3.76E+02 | 1.79E+02 | 9.85E+01 | 6.57E+01 | 1.51E+02 | **1.64E+00** | 8.79E+00 |
| −/≈/+ Nos. | | 16/2/2 | 13/1/6 | 12/1/7 | 13/1/6 | 19/1/0 | 10/2/8 | 12/1/7 | − |
| Ranking | | 4.850 | 4.725 | 4.600 | 4.125 | 5.950 | 4.275 | 4.475 | 3.000 |

"−", "≈" and "+" denote that the performance of the corresponding algorithm is worse than, similar to and better than that of EDC, respectively.

Table 5 reports the experimental results obtained by these eight algorithms, where the results of MA-SW-Chains, MOS, DSPLSO, Rm-ES, MMO-CC and NPDC are directly taken from their corresponding papers, and those of RP-EDA are obtained by running the source code provided by its authors. Table 5 also presents the difference between the fitness means achieved by EDC and each of its competitors according to Cohen's $d$ effect size [55], which is a popular statistical measure independent of the sample size. Generally, no significant difference is considered if an effect size is smaller than 0.2 [55]. Based on Cohen's $d$ effect size, Table 5 labels a result with a symbol "−", "≈" or "+", which indicates that the corresponding result is worse than, similar to or better than the result achieved by EDC, respectively.

The results in Table 5 demonstrate that EDC outperforms its seven competitors on most functions and exhibits the best overall performance. It performs no worse than MA-SW-Chains, MOS, DSPLSO and Rm-ES on 18, 14, 13 and 14 out of total 20 functions, respectively. Compared with RP-EDA, EDC gets better solutions for all the functions except CEC'2010$_3$, where the two algorithms both achieve the global optimum. The comparison between the results in columns four and five in Tables 2-4 reveals that although RP-EDA is not so good at medium-scale problems, it has a definite edge over the optimizer of EDC, i.e., GSM-GEDA, on LSOPs. Then it can be concluded that the superiority of EDA over its non-DC-based competitors mainly stems from its DC strategy, which greatly reduces the search space of an LSOP by properly dividing it into several much smaller subspaces.

When it comes to the two DC-based algorithms, EDC also shows very competitive performance. It surpasses MMO-CC and NPDC on 10 and 12 functions, and yields similar solutions for the other 2 and 1 functions, respectively. The main advantage of EDC over MMO-CC and NPDC consists in that it can easily generate a desirable decomposition by conducting operations in the eigenspace instead of the original solution space.

A closer observation on Table 5 further verifies the conclusion drawn in subsection 4.2, i.e., EDC generally exhibits excellent performance on partially separable and nonseparable functions (such as CEC'2010$_5$, CEC'2010$_{12}$ and CEC'2010$_{19}$), but is not so competitive on separable ones (such as CEC'2010$_1$). For instance, it achieves a better solution than each of its seven competitors on the nonseparable function CEC'2010$_{19}$ by at least four orders of magnitude. With regard to separable functions, EDC is definitely defeated by MA-SW-Chains, MOS, DSPLSO, MMO-CC and NPDC on CEC'2010$_1$, although it yields a much better solution than Rm-ES and RP-EDA for this function and finds the global optimum for CEC'2010$_3$.

To further measure the performance differences between EDC and the other seven algorithms, the last row of Table 5 reports their rankings obtained through the Friedman test. It is evident that EDC can be ranked first with an absolute advantage. This observation is consistent with the comparison results obtained with Cohen'$d$ effect size.

**4.4 Effectiveness of the space transformation technique**

To delve into the effectiveness of the space transformation technique in EDC, we specially designed a variant of EDC by removing its space transformation operations. This means that the new variant conducts decomposition in the original solution space, and so is named ODC. It can be easily implemented by removing steps 7-9 in Algorithm 1. We tested EDC and ODC on 19 functions, including CEC'2005$_1$-CEC'2005$_{14}$ with 200 dimensions, CEC'2010$_1$-CEC'2010$_3$ and CEC'2010$_{19}$-CEC'2010$_{20}$ with 1000 dimensions. The partially separable functions CEC'2010$_4$-CEC'2010$_{18}$ were excluded from this experiment because they involve both separable and nonseparable subcomponents, on which it is inconvenient to analyze the effectiveness of the space transformation technique.

Table 6 reports the experimental results of EDC and ODC, where "−", "≈" and "+" indicate that the performance of ODC is worse than, similar to and better than that of EDC, respectively, according to the Wilcoxon's rank sum test at a 0.05 significance level. The best results in Table 6 are marked in bold. From Table 6, the following observations can be obtained:

Table 6. The mean values and standard deviations (mean±standard deviation) of the FEVs obtained by EDC and ODC over 25 independent runs on 19 functions. Statistical results are obtained based on Wilcoxon's rank sum test at a significance level of 0.05.

| Function | Dimension | Modality | Separability | ODC | EDC |
|---|---|---|---|---|---|
| CEC'2005$_1$ | 200 | unimodal | separable | **0.00E+00±0.00E+00** ≈ | **0.00E+00±0.00E+00** |
| CEC'2005$_2$ | 200 | unimodal | nonseparable | 2.04E+04±2.36E+03 − | **0.00E+00±0.00E+00** |
| CEC'2005$_3$ | 200 | unimodal | nonseparable | 4.30E+06±7.54E+05 − | **1.79E+02±1.40E+02** |
| CEC'2005$_4$ | 200 | unimodal | nonseparable | 1.02E+05±1.67E+04 − | **2.18E+00±7.51E+00** |
| CEC'2005$_5$ | 200 | unimodal | nonseparable | 1.04E+04±1.19E+03 − | **3.18E+02±2.80E+02** |
| CEC'2005$_6$ | 200 | multimodal | nonseparable | **1.17E+02±1.26E+01** − | 1.29E+02±2.63E+01 |
| CEC'2005$_7$ | 200 | multimodal | nonseparable | 1.85E−02±4.43E−02 ≈ | **4.63E−03±2.31E−02** |
| CEC'2005$_8$ | 200 | multimodal | nonseparable | **2.14E+01±1.31E−02** ≈ | 2.14E+01±1.68E−02 |
| CEC'2005$_9$ | 200 | multimodal | separable | **4.33E+01±5.72E+00** + | 6.89E+01±9.89E+00 |
| CEC'2005$_{10}$ | 200 | multimodal | nonseparable | 5.84E+01±7.09E+00 − | **5.32E+01±8.35E+00** |
| CEC'2005$_{11}$ | 200 | multimodal | nonseparable | 1.24E+01±4.25E+00 − | **6.55E+00±5.06E+00** |
| CEC'2005$_{12}$ | 200 | multimodal | nonseparable | 9.38E+04±4.76E+04 ≈ | **8.05E+04±5.36E+04** |
| CEC'2005$_{13}$ | 200 | multimodal | nonseparable | 2.34E+01±1.28E+00 − | **2.11E+01±1.47E+00** |
| CEC'2005$_{14}$ | 200 | multimodal | nonseparable | **9.48E+01±3.88E−01** ≈ | 9.48E+01±3.69E−01 |
| CEC'2010$_1$ | 1000 | unimodal | separable | **0.00E+00±0.00E+00** + | 3.18E+06±2.96E+05 |
| CEC'2010$_2$ | 1000 | multimodal | separable | **3.19E+02±1.23E+01** + | 5.62E+02±4.50E+01 |
| CEC'2010$_3$ | 1000 | multimodal | separable | **0.00E+00±0.00E+00** ≈ | 0.00E+00±0.00E+00 |
| CEC'2010$_{19}$ | 1000 | unimodal | nonseparable | 2.25E+06±7.88E+04 − | **5.27E+00±3.08E+00** |
| CEC'2010$_{20}$ | 1000 | multimodal | nonseparable | 9.39E+02±2.97E−01 − | **9.34E+02±8.79E+00** |
| −/≈/+ Nos. | - | - | - | 10/6/3 | - |

"−", "≈" and "+" denote that the performance of ODC is worse than, similar to and better than that of EDC, respectively.

1) The space transformation technique helps EDC outperform ODC on most functions. The last row of Table 6 summarizes that EDC performs no worse than ODC on 16 out of 19 functions, and is only surpassed by ODC on 3 functions. This means that the space transformation technique in EDC contributes a lot to its excellent performance since it is the only difference between EDC and ODC.

2) The space transformation technique makes EDC good at tacking nonseparable functions but not so competitive on separable functions. A closer observation on Table 6 indicates that EDC performs no worse than ODC on all the 14 nonseparable functions. As for the five separable functions, EDC achieves the same results with ODC on two functions (CEC'2005$_1$ and CEC'2010$_3$), but is defeated by ODC on the other three ones. This is understandable because the space transformation operation can significantly weaken the strong dependencies among variables in nonseparable functions, but is nonessential for separable functions, whose variables are all independent with each other in the original solution space.

3) The space transformation technique shows different performance on functions of different modalities. As indicated above, the space transformation technique endows EDC superiority over ODC on nonseparable functions. However, this kind of superiority is highly significant on unimodal functions, but weakens to some extent on multimodal ones. This conclusion can be supported by the observation that EDC achieves better solutions than ODC for unimodal functions, such as CEC'2005$_2$-CEC'2005$_5$ and CEC'2010$_{19}$, by several orders of magnitudes, while most of the better solutions it produces for multimodal functions are in the same order of magnitude with the corresponding ones yielded by ODC. The reason is twofold. On the one hand, it is intrinsically more difficult for an optimizer to find desirable solutions for a multimodal functions since it may have many abundant local optima. On the other hand, the complicated landscape of a multimodal function, which generally involves many basins of attraction, makes it much harder to learn an accurate enough eigenspace through a computing method such as SVD. In spite of this, EDC still improves ODC on nonseparable multimodal functions, which mainly profits from the effectiveness of its space transformation technique.

Fig. 6 presents the evolution curves of EDC and ODC, where the two functions CEC'2010$_3$ and CEC'2010$_{19}$ are taken as examples. It can be observed from Fig. 6(a) that for the separable function CEC'2010$_3$, EDC and ODC generate almost the same evolution curve as well as the same final result, which reveals that the space transformation operation in EDC hardly affects its convergence speed on this function. As for the nonseparable function CEC'2010$_{19}$, EDC keeps a desirable improvement tendency during the evolution process and finally achieves a much better solution, whereas ODC gets stuck quickly.

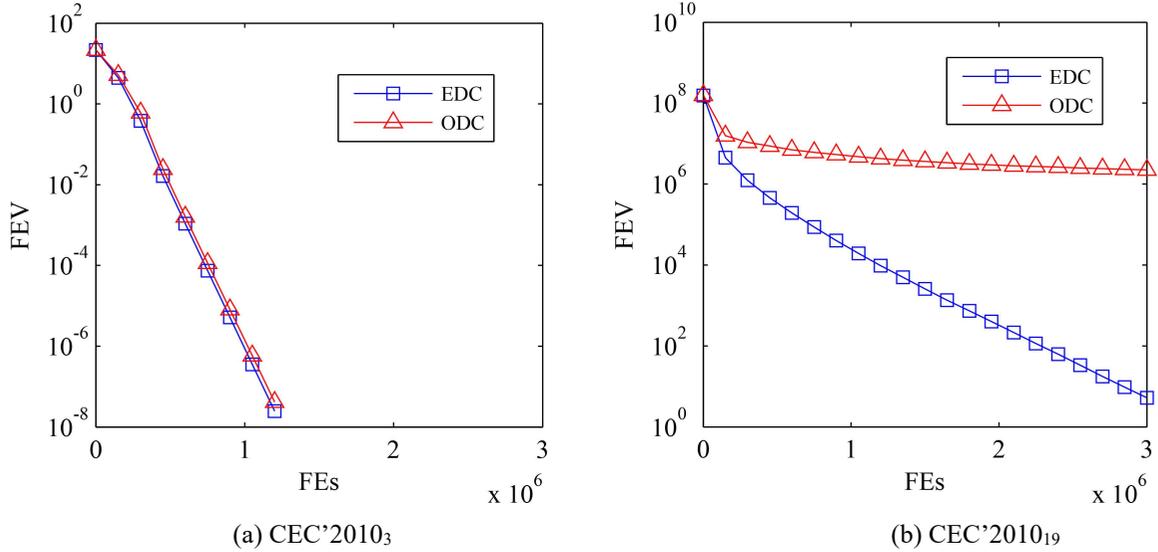

(a) CEC'2010$_3$          (b) CEC'2010$_{19}$

Figure 6. The evolution curves of EDC and ODC on two functions.

## 5. Conclusions and future work

This study first investigates the reasons for the inefficiency of existing DC-based methods and points out that the direct decomposition in the original solution space may be a main reason, because this approach either neglects possible strong variable dependencies or asks for lots of computation resources. Based on this finding, we suggest an eigenspace divide-and-conquer approach for large-scale optimization. By taking advantages of a space transformation technique and the DC strategy, EDC demonstrates great competitiveness in solving complicated LSOPs that are recognized to be hard for existing DC-based algorithms. The sensitivity test indicates that EDC is rather robust to its parameters and has good scalability to problem dimension. The numerical comparison with several state-of-the-art algorithms, including DC-based algorithms and non-DC-based ones, demonstrates that EDC performs best on two well-known benchmark suites. Further experimental study reveals that the space transformation operation in EDC is very effective and contributes a lot to the excellent performance of EDC.

EDC provides a new potential way to address LSOPs and leaves much room to improve its performance. Firstly, it is beneficial to attempt some other space transformation techniques besides SVD such that more proper eigenspaces can be learnt. Secondly, instead of equally tackling different eigenvariables, it is more reasonable to pay more computation effort to dominant eigenvariables. Finally, it is also necessary to improve EDC by introducing more efficient decomposers and optimizers into it.